\documentclass[11pt,a4paper]{article}
\usepackage[hyperref]{acl2020}
\usepackage{times}
\usepackage{latexsym}

\usepackage{url}
\usepackage{graphicx}
\usepackage{subcaption}
\usepackage{tikz}
\usepackage{amsfonts}
\usepackage{amsmath}
\usepackage{paralist}
\usepackage{xspace}
\usepackage{afterpage}
\usepackage{booktabs}
\usepackage{multirow}

\usepackage{rotating}

\newcommand{\ibs}{{\sf MIBs}\xspace}
\newcommand{\mlb}{{\sf MLBs}\xspace}
\newcommand{\intrinsicM}{{\sf inBias}\xspace}

\usepackage{array}
\graphicspath{figs}
\usepackage{diagbox}

\usepackage{adjustbox}

\usepackage{microtype}

\aclfinalcopy 
\def\aclpaperid{1209} 

\title{Gender Bias in Multilingual Embeddings and Cross-Lingual Transfer}

\author{Jieyu Zhao$^\S$ \thanks{\hspace{1.5mm}Most of the work was done while the first author was an intern at Microsoft Research.} \qquad
 Subhabrata Mukherjee$^\ddag$ \qquad 
 Saghar Hosseini$^\ddag$   \\
{\bf Kai-Wei Chang$^\S$ \qquad
 Ahmed Hassan Awadallah$^\ddag$}
\\
  $^\S$University of California, Los Angeles  \qquad$^\ddag$Microsoft Research AI  \\ 
  \{jyzhao, kwchang\}@cs.ucla.edu \\
  \{Subhabrata.Mukherjee, Saghar.Hosseini, hassanam\}@microsoft.com
}

\date{}

\begin{document}
\maketitle
\begin{abstract}
Multilingual representations embed words from many languages into a single semantic space such that words with similar meanings are close to each other regardless of the language. These embeddings have been widely used in various settings, such as cross-lingual transfer, where a natural language processing (NLP) model trained on one language is deployed to another language. 
While the cross-lingual transfer techniques are powerful, they carry gender bias from the source to target languages. 
In this paper, we study gender bias in multilingual embeddings and how it affects transfer learning for NLP applications. We create a multilingual dataset for bias analysis and propose several ways for quantifying bias in multilingual representations from both the intrinsic and extrinsic perspectives. Experimental results show that the magnitude of bias in the multilingual representations changes differently when we align the embeddings to different target spaces and that the alignment direction can also have an influence on the bias in  transfer learning.  We further provide recommendations for using the multilingual word representations for downstream tasks.
\end{abstract}

\section{Introduction}
\label{sec:intro}
Natural Language Processing (NLP) plays a vital role in applications used in our daily lives.
Despite the great performance inspired by the advanced machine learning techniques and large available datasets, there are potential societal biases embedded in these NLP tasks -- where the systems learn inappropriate correlations between the final predictions and sensitive attributes such as gender and race. For example, \citet{zhao2018gender} and \citet{rachel18} demonstrate that coreference resolution  systems perform unequally on different gender groups. 
Other studies show that such bias is exhibited in various components of the NLP systems, such as the training dataset~\cite{zhao2018gender, rachel18}, the embeddings~\cite{bolukbasi2016man, caliskan2017semantics,zhou2019grammaticalgenderbias,manzini2019black} as well as the pre-trained models~\cite{zhao2019gender,kurita2019quantifying}.

Recent advances in NLP require large amounts of training data.
Such data may be available for resource-rich languages such as English, but they are typically absent for many other languages. 
Multilingual word embeddings align the  embeddings from various languages to the same shared embedding space which enables  transfer learning by training the model in one language and adopting it for another one~\cite{ammar2016massively,ahmad2019crosslingual,meng2019target, chen2019multisource}.
Previous work has proposed different methods to create multilingual word embeddings. One common way is to first train the monolingual word embeddings separately and then align them to the same space~\cite{conneau2017word, joulin2018loss}.
While multiple efforts have focused on improving the models' performance on low-resource languages, less attention is given to understanding the bias in  cross-lingual transfer learning settings.

 In this work, we aim to understand the bias in multilingual word embeddings. 
In contrast to existing literature that mostly focuses on English, we conduct analyses in multilingual settings.
 We argue that the bias in multilingual word embeddings can be very different from that in English. One reason is that each language has its own properties. For example, in English, most nouns do not have  grammatical gender, while in Spanish, all nouns do. Second, when we do the alignment to get the multilingual word embeddings, the choice of target space may cause  bias.  Third,  when we do transfer learning based on multilingual word embeddings,  the alignment methods, as well as the transfer procedure can potentially influence the bias in downstream tasks.
Our experiments confirm that bias exists in the multilingual embeddings and such bias also impacts the cross-lingual transfer learning tasks. We observe that the transfer  model based on the multilingual word embeddings shows discrimination against genders. To discern such bias, we perform analysis from both the corpus and the embedding perspectives, showing that both contribute to the bias in transfer learning.
Our contributions are summarized as follows: 
\begin{compactitem}
    \item We build datasets for studying the gender bias in multilingual NLP systems.\footnote{Code and data will be available at \url{ https://aka.ms/MultilingualBias}.}
    \item We analyze  gender bias in multilingual word embeddings from both intrinsic and extrinsic perspectives. Experimental results show that the pre-trained monolingual word embeddings, the alignment method as well as the transfer learning can have an impact on the gender bias.
    \item We show that simple mitigation methods can help to reduce the bias in multilingual word embeddings and discuss directions for future work to further study the problem. We provide several recommendations for bias mitigation in cross-lingual transfer learning.
\end{compactitem}

\section{Related Work}
\label{sec:related_work}
\paragraph{Gender Bias in Word Representations}
Word embeddings are widely used in different NLP applications. They represent words using  low dimensional vectors. \citet{bolukbasi2016man} find that, in the embedding space, occupation words such as ``professor'' and ``nurse'' show discrepancy concerning the genders.  Similarly, \citet{caliskan2017semantics} also reveal the gender stereotypes in the English word embeddings based on the Word Embedding Association Test (WEAT). However, both works only consider English and cannot be directly adapted to other languages such as Spanish. \citet{mccurdy2017grammatical} reveal that bias exists in languages with grammatical gender while  \citet{zhou2019grammaticalgenderbias} and \citet{lauscher2019we} show that there is bias in bilingual word embeddings. However, none of them  consider the cross-lingual transfer learning which is an important application of the multilingual word embeddings. 
To mitigate the bias in word embeddings, various approaches have been proposed~\cite{bolukbasi2016man,zhao2018learning}. In contrast to these methods in English embedding space, we propose to mitigate the bias from the multilingual perspectives. Comparing to \citet{zhou2019grammaticalgenderbias}, we show that a different choice of alignment target can help to reduce the bias in multilingual embeddings from both intrinsic and extrinsic perspectives.

\paragraph{Multilingual Word Embeddings and Cross-lingual Transfer Learning}
Multilingual word embeddings represent words from different languages using the same embedding space which  enables cross-lingual transfer learning~\cite{ruder2017survey}. The model is trained on a labeled data rich language and adopted to another language where no or a small portion of labeled data is available~\cite{duong2015low, guo2016representation}. To get the multilingual word embeddings, \citet{mikolov2013exploiting} learn a linear mapping between the source and target language.  
However, \citet{xing2015normalized} argue that there are some inconsistencies in directly learning the linear mapping. To solve those limitations, they constrain the embeddings to be normalized and enforce an orthogonal transformation. 
While those methods  achieve reasonable results on benchmark datasets, they all suffer from the hubness problem which is solved by adding cross-domain  similarity constraints~\cite{conneau2017word,joulin2018loss}.
Our work is based on the multilingual word embeddings achieved by~\newcite{joulin2018loss}. Besides the commonly used multilingual word embeddings obtained by aligning all the embeddings to the English space, we also analyze the embeddings aligned to  different target spaces.

\paragraph{Bias in Other Applications} Besides the bias in word embeddings, such issues have also been demonstrated in other applications, including named entity recognition~\cite{mehrabi2019man}, sentiment analysis~\cite{kiritchenko2018examining}, and natural language inferences~\cite{rudinger2017social}. However, those analyses are limited to English corpus and lack the insight of multilingual situations.

\section{Intrinsic Bias Quantification and Mitigation}
\label{sec:intrinsic}
In this section, we  analyze the gender bias in multilingual word embeddings. Due to the limitations of the available resources in other languages, we analyze the bias in English, Spanish, German and French. However, our systematic evaluation approach can be easily extended to other languages. We first define an evaluation metric for quantifying gender bias in multilingual word embeddings. Note that in this work, we focus on analyzing gender bias from the perspective of occupations.  We then show that when we change the target alignment space, the bias in multilingual word embeddings also changes. Such observations provide us a way to mitigate the bias in multilingual word embeddings -- by choosing an appropriate target alignment space. 

\subsection{Quantifying Bias in Multilingual Embeddings}
We begin with describing \intrinsicM, our proposed evaluation metric for quantifying intrinsic  bias in multilingual word embeddings from  word-level perspective. We then introduce the dataset we collected for quantifying bias in different languages.

\paragraph{Bias Definition}
Given a set of masculine and feminine words, we define \intrinsicM as:
\begin{equation}
\label{eq:bias}
   \text{\intrinsicM} = \frac{1}{N}\sum_{i=1}^{N}|dis(O_{M_i}, S_M) - dis(O_{F_i}, S_F)|,
\end{equation}
where
\begin{equation*}
 dis(O_{G_i}, S) = \frac{1}{|S|}\sum_{s\in S}(1 - \cos{(O_{G_i}, s))}.
   \end{equation*}
Here ($O_{M_i}$, $O_{F_i}$) stands for the masculine and feminine format of the $i$-th occupation word, such as (``doctor'', ``doctora''). $S_M$ and $S_F$ are a set of  gender seed words  that contain male and female gender information in the definitions such as ``he'' or ``she''.

Intuitively, given a pair of masculine and feminine words describing an occupation, such as the words ``\textit{doctor}'' (Spanish, masculine doctor) and ``\textit{doctora}'' (Spanish, feminine doctor), the only difference lies in the gender information. As a result, they should have similar correlations to the corresponding gender seed words such as ``\textit{\'{e}l}'' (Spanish, he) and ``\textit{ella}''  (Spanish, she). If there is a gap between the distance of occupations and corresponding gender, (i.e., the distance between ``doctor'' and ``\'{e}l'' against the distance between ``doctora'' and ``ella''), it means such occupation shows discrimination against gender. 
Note that such metric can also be generalized to other languages without grammatical gender, such as English, by just using the same format of the occupation words. It is also worth noting that our metric is general and can be used to define other types of bias with slight modifications. For example, it can be used to detect age or race bias by providing corresponding seed words (e.g., ``young'' - ``old'' or names correlated with different races).
In this paper we focus on gender bias as the focus of study.
 We  provide detailed descriptions of those words in the dataset collection subsection.

Unlike previous work~\cite{bolukbasi2016man} which requires calculating a gender direction by doing dimensionality reduction, we do not require such a step and hence we can keep all the information in the embeddings. 
The goal of \intrinsicM is aligned to that of WEAT \cite{caliskan2017semantics}. It calculates the difference of targets (occupations in our case) corresponding to different attributes (gender). We use paired occupations in each language, reducing the influence of grammatical gender.
Compared to \citet{zhou2019grammaticalgenderbias}, we do not need to separately generate the two gender directions, as in our definition, the difference of the distance already contains such information. In addition, we no longer need to collect the gender neutral word list. In multilingual settings, due to  different gender assignments to each word (e.g., ``spoon'' is masculine is DE but feminine in ES), it is expensive to collect such resources which can be alleviated by the \intrinsicM metric.

\begin{figure*}[!t]
    \centering
    \begin{subfigure}[b]{0.31\textwidth}
        \includegraphics[width=1.1\textwidth]{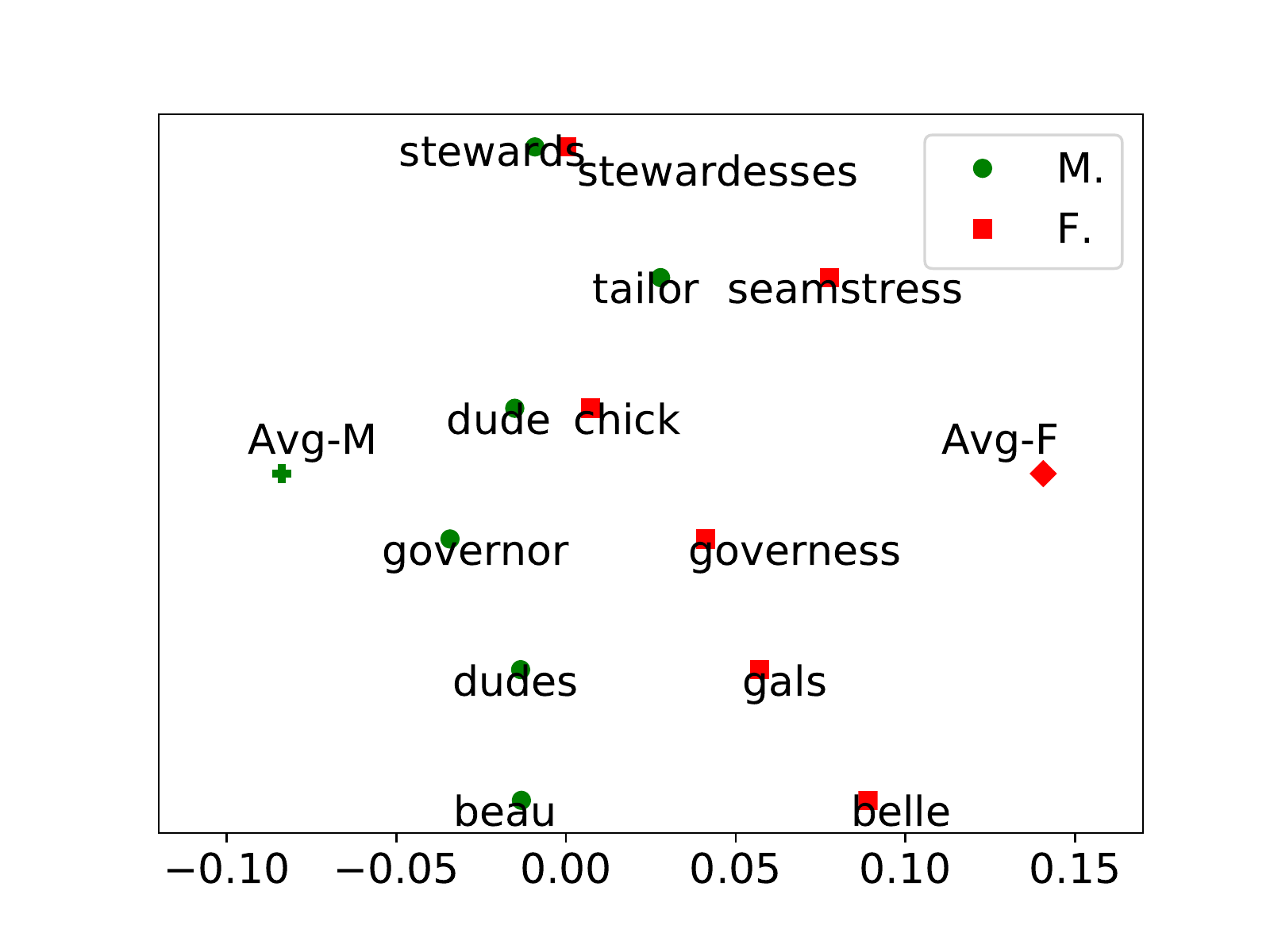}
        \caption{
        Original  es embeddings.
        }
        \label{fig:es_ori_occ}
    \end{subfigure}
    ~
    \begin{subfigure}[b]{0.31\textwidth}
        \includegraphics[width=1.1\textwidth]{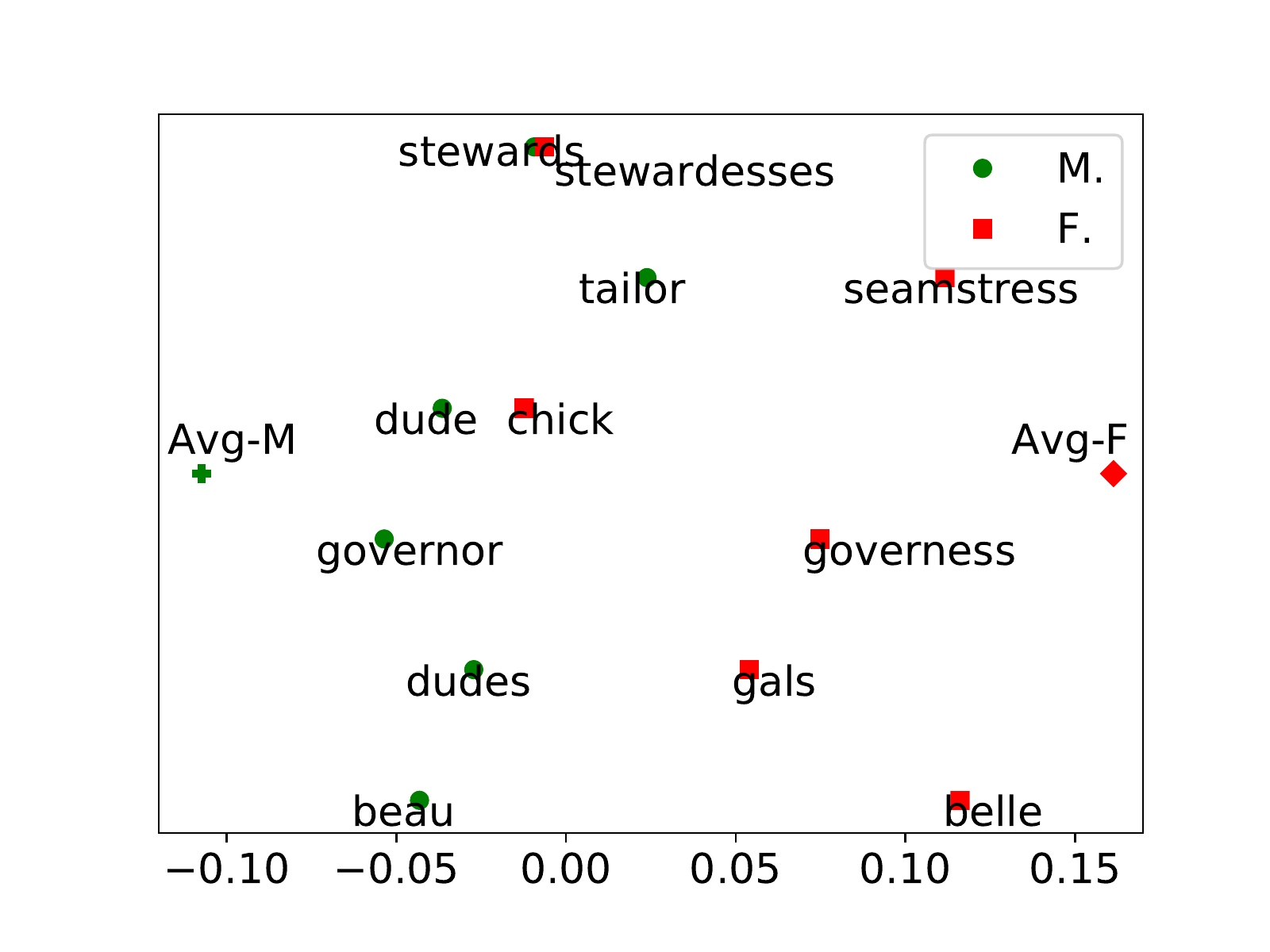}
        \caption{
        In  es-en embeddings.
        }
        \label{fig:es_ali_occ}
    \end{subfigure}
        ~
    \begin{subfigure}[b]{0.31\textwidth}
        \includegraphics[width=1.1\textwidth]{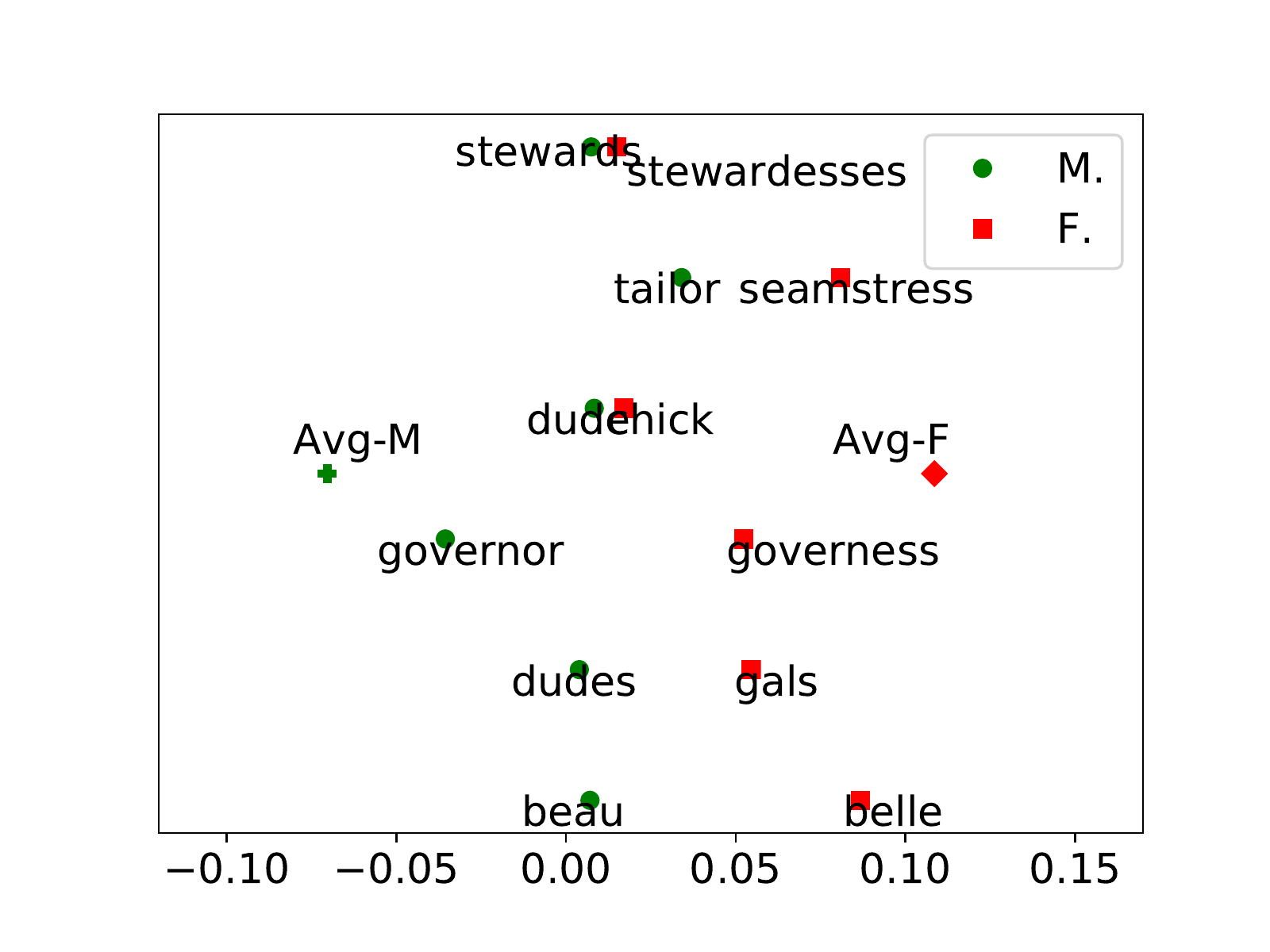}
        \caption{
        In es-de embeddings.
        }
        \label{fig:es_de_occ}
    \end{subfigure}
    \caption{Most biased occupations in ES projected to the gender subspace defined by the difference between two gendered seed words. Green dots are masculine (M.) occupations while the red squares are feminine (F.) ones. We also show the average projections of the gender seed words for male and female genders denoted by ``Avg-M'' and ``Avg-F''.  Compared  to EN, aligning to DE makes the distance between the occupation word and corresponding gender more symmetric.}
    \label{fig:es_plot}
\end{figure*}

\paragraph{Multilingual Intrinsic Bias Dataset}

To conduct the intrinsic bias analysis, we create the \ibs dataset by manually collecting pairs of occupation words and gender seed words in four languages: English (EN), Spanish (ES), German (DE) and French (FR). We choose these four languages as they come from different language families (EN and DE belong to the Germanic language family while ES and FR belong to the Italic language family) and exhibit different gender properties (e.g., in ES, FR and DE, there is grammatical gender).\footnote{We also do analyses with Turkish where there is no grammatical gender and no gendered pronoun. Details are in Sec.~\ref{sec:languageofstudy}.} We refer to languages with grammatical gender as \textsc{gender-rich} languages; and otherwise, as \textsc{gender-less} languages. Among these three gender-rich languages, ES and FR only have feminine and masculine genders while in DE, there is also a neutral gender.
We obtain the feminine and masculine words in EN from~\newcite{zhao2018learning} and extend them by manually adding other common occupations. The English gender seed words are from~\citet{bolukbasi2016man}.  For all the other languages, we  get the corresponding masculine and feminine terms by using online translation systems, such as Google Translate.
We refer to the words that have both masculine and feminine formats in EN (e.g., ``waiter'' and ``waitress'') as \textit{strong gendered} words while others like ``doctor'' or ``teacher'' as \textit{weak gendered} words. In total, there are 257 pairs of occupations  and 10 pairs of gender seed words for each language. 
 In the gender-rich languages, if the occupation only has one lexical format, (e.g., ``prosecutor'' in ES only has the format ``fiscal''), we  add it to both the feminine and the masculine lists.  

\subsection{Characterizing Bias in Multilingual Embeddings}
\label{sec:multi_emb}
As mentioned in Sec.~\ref{sec:intro}, multilingual word embeddings can be generated by first training word embeddings for different languages individually and then aligning those embeddings to the same space. During the alignment, one language is chosen as target  and the embeddings from other languages are projected onto this target space.
We conduct comprehensive analyses on the \ibs dataset to understand: 1) how gender bias exhibits in embeddings of different languages; 2) how the alignment target affects the gender bias in the embedding space; and 3) how the quality of multilingual  embeddings is affected by  choice of the target language.

For the monolingual embeddings of individual languages and the multilingual embeddings that used English as the target language (*-en),\footnote{We  refer to the aligned multilingual word embeddings using the format src-tgt. For example, ``es-en'' means we align the ES embeddings to the EN space. An embedding not following such format refers to a monolingual embedding.} we use the publicly available  fastText embeddings trained on 294 languages in Wikipedia~\cite{bojanowski2017enriching,joulin2018loss}. For all  other embeddings  aligned to a target space other than EN, we adopt the RCSLS alignment model~\cite{joulin2018loss} based on the same hyperparameter setting (details are in Appendix).

\begin{table}[!t]
\small
\centering
    \begin{tabular}{|c|c|c|c|c|}
    \hline
         \multirow{2}{*}{Source} & \multicolumn{4}{c|}{Target} \\
          \cline{2-5}
           & EN & ES & DE & FR \\
           \hline
           EN &  \textbf{0.0830} & 0.0639* &  0.0699* & 0.0628* \\
           \hline
           ES & 0.0889* & \textbf{0.0803} & 0.0634* &  0.0642*\\
           \hline
           DE & 0.1124 &  0.0716* & \textbf{0.1079} & 0.0805* \\
           \hline
           FR & 0.1027 & 0.0768* & 0.0782* &  \textbf{0.0940} \\
           \hline
          
    \end{tabular}
    \caption{\intrinsicM~score before and after alignment to different target spaces. Rows stands for the source languages while columns are the target languages. The diagonal values stand for the bias in the original monolingual word embeddings. Here * indicates the difference between the bias before and after alignment is statistically significant ($p<0.05$). } 
    \label{tab:bias_res}
\end{table}

\subsubsection{Analyzing Bias  before Alignment}
We examine the bias using  four languages mentioned previously based on all the word pairs in the ~\ibs. Table~\ref{tab:bias_res} reports the \intrinsicM score on this dataset. The diagonal values here stand for the bias in each language before alignment. Bias commonly exists across all the four languages. Such results are also supported by WEAT  in ~\citet{zhou2019grammaticalgenderbias}, demonstrating the validity of our metric. What is more, comparing those four languages, we find DE and FR have stronger biases comparing to  EN and ES.

\subsubsection{How will the bias change when aligned to different languages?}
Commonly used multilingual word embeddings  align all languages to the English space. However, our analysis shows that the bias in the multilingual word embeddings can change if we choose a different target space.
 All the results are shown in Table~\ref{tab:bias_res}. 
 Specifically, when we align the embeddings to the gender-rich languages, the bias score will be lower compared to that in the original embedding space.  In the other situation, when aligning the embeddings to the  gender-less language space (i.e., EN in our case), the bias  increases. For example, in original EN, the bias score is $0.0830$ and when we align EN to ES, the bias decreases to $0.0639$ with $23\%$ reduction in the bias score. However,  the bias in ES embeddings  increases to $0.0889$ when aligned to EN while only $0.0634$ when aligned to DE.\footnote{We show the bias for all the 257 pairs of words in EN. In the appendix, we also show the bias for strong gendered words and weak gendered words separately.}
In Fig.~\ref{fig:es_plot}, we show the examples of word shifting along the gender direction when aligning ES to different languages. The gender direction is calculated by the difference of male gendered seeds and female gendered seeds. 
We observe the feminine occupations are further away from female seed words than masculine ones, causing the resultant bias. 
In comparison to using EN as target space, when aligning ES to DE, the distance between masculine and feminine occupations with corresponding gender seed words become more symmetric, therefore reducing the \intrinsicM score.

\begin{table}[!t]
\small
\centering
\begin{adjustbox}{max width=\textwidth}
    \begin{tabular}{|c|c|c|c|c|}\hline
         \multirow{2}{*}{Source} & \multicolumn{4}{c|}{Target} \\
          \cline{2-5}
          & EN & ES & DE & FR \\
           \hline
           EN & -  & 83.08 &  78.60  & 83.00  \\
           \hline
           ES & 86.40 & - & 72.40 &  87.27 \\
           \hline
           DE & 76.33 &  69.80  & - & 78.13  \\
           \hline
           FR & 84.27 & 84.80  & 75.53 & - \\
           \hline
    \end{tabular}
    \end{adjustbox}
    \caption{Performance (accuracy \%) of the BLI task for the aligned embeddings. Row stands for the source language and column is the target language. The values in the first row are from~\citet{joulin2018loss}.}
    \label{tab:acc_res}
\end{table}

\paragraph{What words changed most after the alignment?} We are  interested in understanding how the gender bias of words changes  after we do the alignment. To do this, we look at the top-15 most and least changed words. We find that  in each language, the strongest bias comes from the strong gendered words; while the least bias happens among weak gendered words.
 When we align EN embeddings to gender-rich languages, bias in the strong gendered words will change most significantly; and the weak gendered words will change least significantly. When we align gender-rich languages to EN, we observe a similar trend.
Among all the alignment cases, gender seed words used in Eq.~\eqref{eq:bias} do not change significantly.

\subsubsection{Bilingual Lexicon Induction}
To evaluate the quality of word embeddings after the alignment, we test them on the bilingual lexicon induction (BLI) task~\cite{conneau2017word}  goal of which is to induce the translation of source words by looking at their nearest neighbors.
We  evaluate the embeddings on the MUSE dataset with the CSLS metric ~\cite{conneau2017word}.  

We conduct  experiments among all the pair-wise alignments of the four languages. The results are shown in Table~\ref{tab:acc_res}. Each row depicts  the source language, while the column depicts the target language. When aligning languages to different target spaces, we do not observe a significant performance difference in comparison to aligning to EN in most cases. This confirms the possibility to use such embeddings in downstream tasks.  However, due to the limitations of available resources,  we only show the result on the four languages and it may change when using different languages.

\subsubsection{Languages of Study}
\label{sec:languageofstudy}
In this paper, we mainly focus on four European languages from different language families,  partly caused by the limitations of the currently available resources. We do a simplified analysis on Turkish  (TR) which belongs to the Turkic language family. In TR, there is no grammatical gender for both nouns and pronouns, i.e., it uses the same pronoun ``o'' to refer to ``he'', ``she'' or ``it''.
The original bias in TR is $0.0719$ and when we align it to EN, the bias remains almost the same at $0.0712$. 
When  aligning EN to TR, we can reduce the intrinsic bias in EN from $0.0830$ to $0.0592$, with $28.7\%$ reduction. However, the BLI task shows that the performance on such aligned embeddings drops significantly: only $53.07\%$ when aligned to TR but around $80\%$ when aligned to the other four languages. Moreover, as mentioned in \citet{ahmad2019difficulties}, some other languages such as Chinese and Japanese cannot align well to English. 
Such situations require more investigations and forming a direction for future work.

\begin{table}[!h]
\small
\centering
\begin{adjustbox}{ max width=\textwidth}
    \begin{tabular}[m]{|c|c|c|c|c|}
        \hline
         \multirow{2}{*}{Source} & \multicolumn{4}{c|}{Target} \\
          \cline{2-5}
          & ENDEB & ES & DE & FR \\
           \hline
           ENDEB & 0.0501*  & 0.0458* & 0.0524*  & 0.0441*  \\
           \hline
           ES & 0.0665* &  0.0803 & - &  - \\
           \hline
           DE & 0.0876* & -  & 0.1079 & -  \\
           \hline
           FR & 0.0905 & -  & - & 0.0940 \\
           \hline
    \end{tabular}
\end{adjustbox}
    \caption{\intrinsicM score before and after alignment to ENDEB. * indicates statistically significant difference between the bias in original and aligned embeddings.}
    \label{tab:bias_res_endeb}
\end{table}

\subsection{Bias after Mitigation} 
Researchers have proposed different approaches to mitigate the bias in EN word embeddings~\cite{bolukbasi2016man,zhao2018learning}. 
Although these approaches cannot entirely remove the  bias~\cite{gonen2019lipstick}, they significantly reduce the bias in English embeddings. We refer to such embedding as  \emph{ENDEB}.
We analyze how the bias changes after we align the embeddings to such ENDEB space. The ENDEB embeddings are obtained by adopting the method in \citet{bolukbasi2016man} on the original fastText monolingual word embeddings. 
Table~\ref{tab:bias_res_endeb} and \ref{tab:_accres_endeb} show the bias score and BLI performance when we do the alignment between ENDEB and other languages. Similar to \citet{zhou2019grammaticalgenderbias}, we find that when we align other embeddings to the ENDEB space, we can reduce the bias in those embeddings. What is more, we  show that we can reduce the bias in ENDEB embeddings further when we align it to a gender-rich language such as ES while keeping the functionality of the embeddings, which is consistent with our previous observation in Table~\ref{tab:bias_res}.  Besides, comparing aligning to gender-rich languages and to ENDEB, the former one can reduce the bias more.

\section{Extrinsic Bias Quantification and Mitigation}
\label{sec:extrinsic}
In addition to the intrinsic bias in multilingual word embeddings, we also analyze the downstream tasks, specifically in the cross-lingual transfer learning. One of the main challenges here is the absence of appropriate datasets. To motivate further research in this direction, we build a new dataset called \mlb. Experiments demonstrate that bias in multilingual word embeddings can also have an effect on models transferred to different languages. We further show how  mitigation  methods can help to reduce the bias in the transfer learning setting. 

\begin{table}[!t]
\small
\centering
\begin{adjustbox}{max width=\textwidth}
    \begin{tabular}{|c|c|c|c|c|}
        \hline
         \multirow{2}{*}{Source} & \multicolumn{4}{c|}{Target} \\
          \cline{2-5}
          & ENDEB & ES & DE & FR \\
           \hline
           ENDEB & - & 84.07 & 79.13  & 83.27  \\
           \hline
               \hline
        \multirow{2}{*}{Target} & \multicolumn{4}{c|}{Source} \\
          \cline{2-5}
          & ENDEB & ES & DE & FR \\
           \hline
           ENDEB & - & 86.07 & 76.27  & 84.33  \\
           \hline
    \end{tabular}
    \end{adjustbox}
    \caption{Performance (accuracy \%) on the BLI task using the aligned embeddings based on ENDEB embeddings. The top one is the result of aligning ENDEB to other languages while the bottom is to align other languages to ENDEB.}
    \label{tab:_accres_endeb}
\end{table}

\begin{table}[!t]
\small
\centering
\begin{adjustbox}{max width=\textwidth}
    \begin{tabular}{|c|c|c|c|c|}
        \hline
        Language &   EN & ES & DE &  FR  
        \\
          \hline
          \#occupation & 28 & 72  & 27 & 27\\
          \hline 
          \#instance & 397,907 & 82,863  & 12,976  & 59,490\\
          \hline
  \end{tabular}
  \end{adjustbox}
    \caption{Statistics of the \mlb for each language. }
    \label{tab:bios_occ_num}
\end{table}

\subsection{Quantifying Bias in Multilingual Models}
In this section, we provide details of the dataset we collected for the extrinsic bias analysis as well as the metric we use for the bias evaluation.
\subsubsection*{Multilingual BiosBias Datasets}
\citet{de2019bias} built an English BiosBias dataset to evaluate the bias in predicting the occupations of people when provided with a short biography on the bio of the person written in third person. To evaluate the bias in cross-lingual transfer settings, we build the Multilingual BiosBias (\mlb) Dataset which contains bios in different languages.

\textit{Dataset Collection Procedure} 
We collect a list of common occupations for each language and follow the data collection procedure used for the English dataset~\cite{de2019bias}. To identify bio paragraphs, we use the pattern  ``NAME is an OCCUPATION-TITLE'' where name is recognized in each language by using the corresponding Named Entity Recognition model from spaCy.\footnote{https://spacy.io/usage/models} To control for the same time period for datasets across languages, we  process the same set of Common Crawl dumps ranging from the year 2014 to 2018. For the occupations, we  use both the feminine and masculine versions of the word in the gender-rich languages. For EN, we use the existing BiosBias dataset.

The number of occupations in each language is shown in Table~\ref{tab:bios_occ_num}.
As the bios are written in third person, similar to ~\citet{de2019bias}, we   extract the binary genders based on the gendered pronouns in each language, such as ``he'' and ``she''. 

\begin{figure*}[!t]
\centering
    \begin{subfigure}[b]{0.23\textwidth}
        \includegraphics[width=1.18\textwidth]{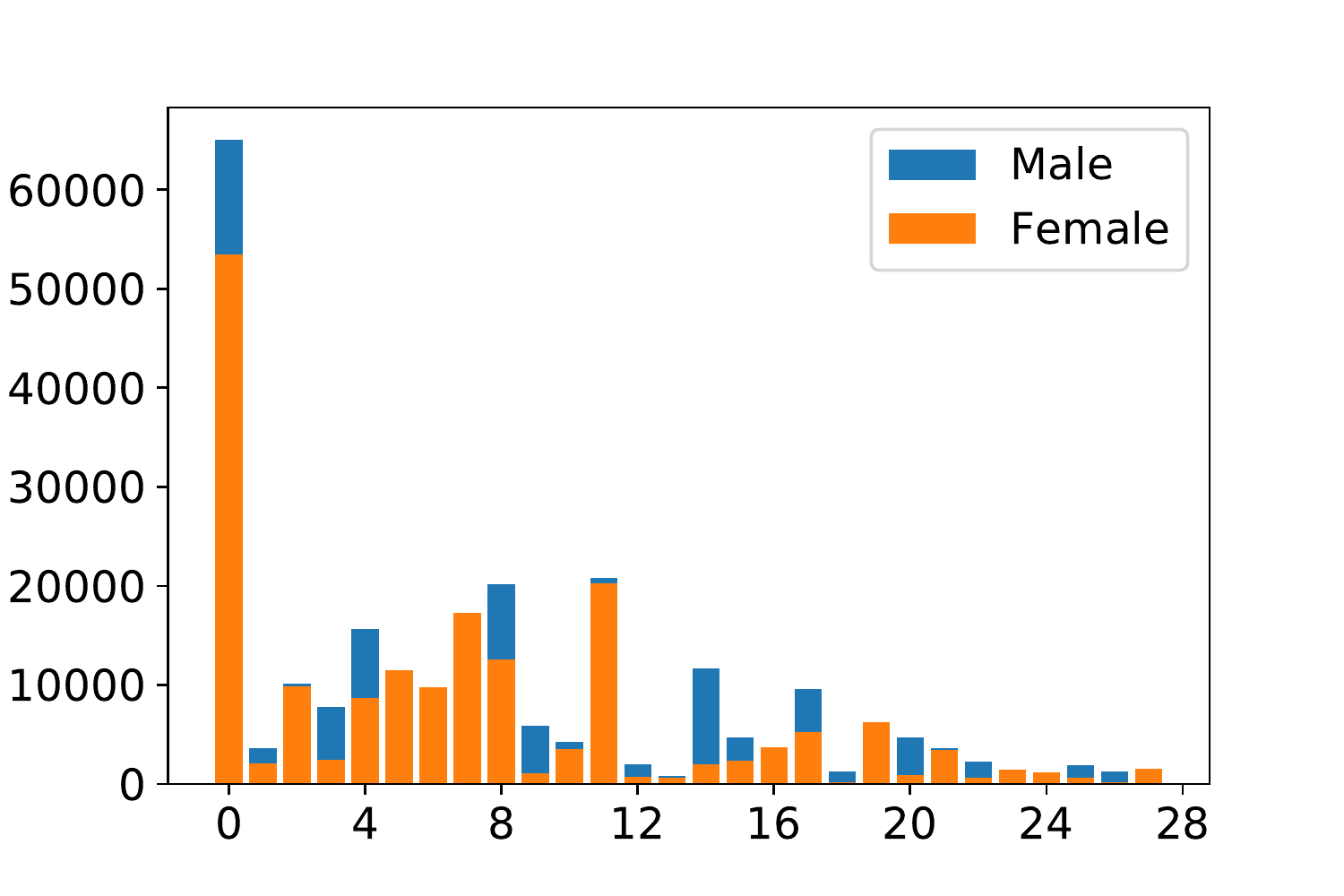}
        \caption{
        EN
        }
        \label{fig:en_stat}
    \end{subfigure}
    ~
    \begin{subfigure}[b]{0.23\textwidth}
        \includegraphics[width=1.18\textwidth]{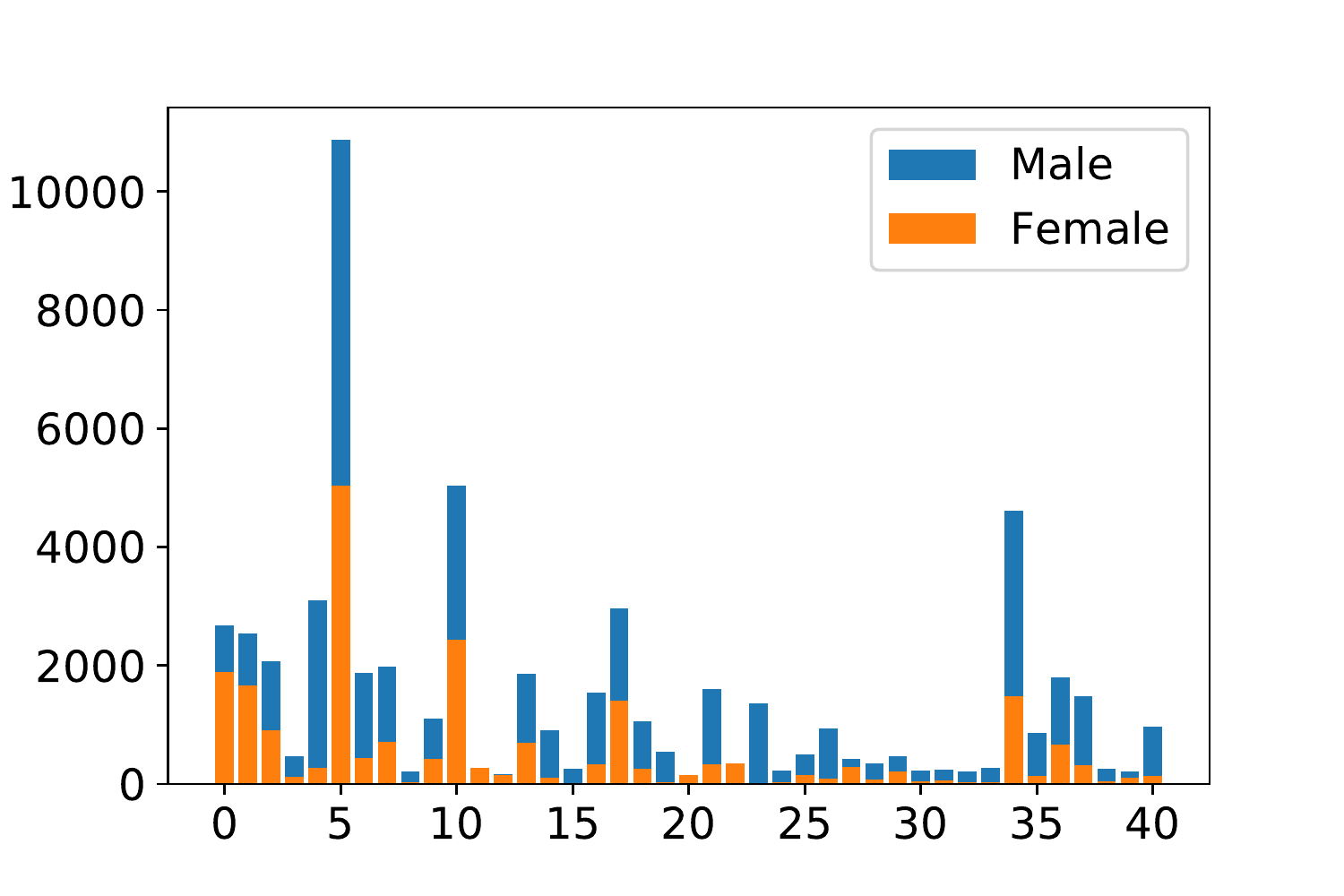}
        \caption{
        ES
        }
        \label{fig:es_stat}
    \end{subfigure}
        ~
    \begin{subfigure}[b]{0.23\textwidth}
        \includegraphics[width=1.18\textwidth]{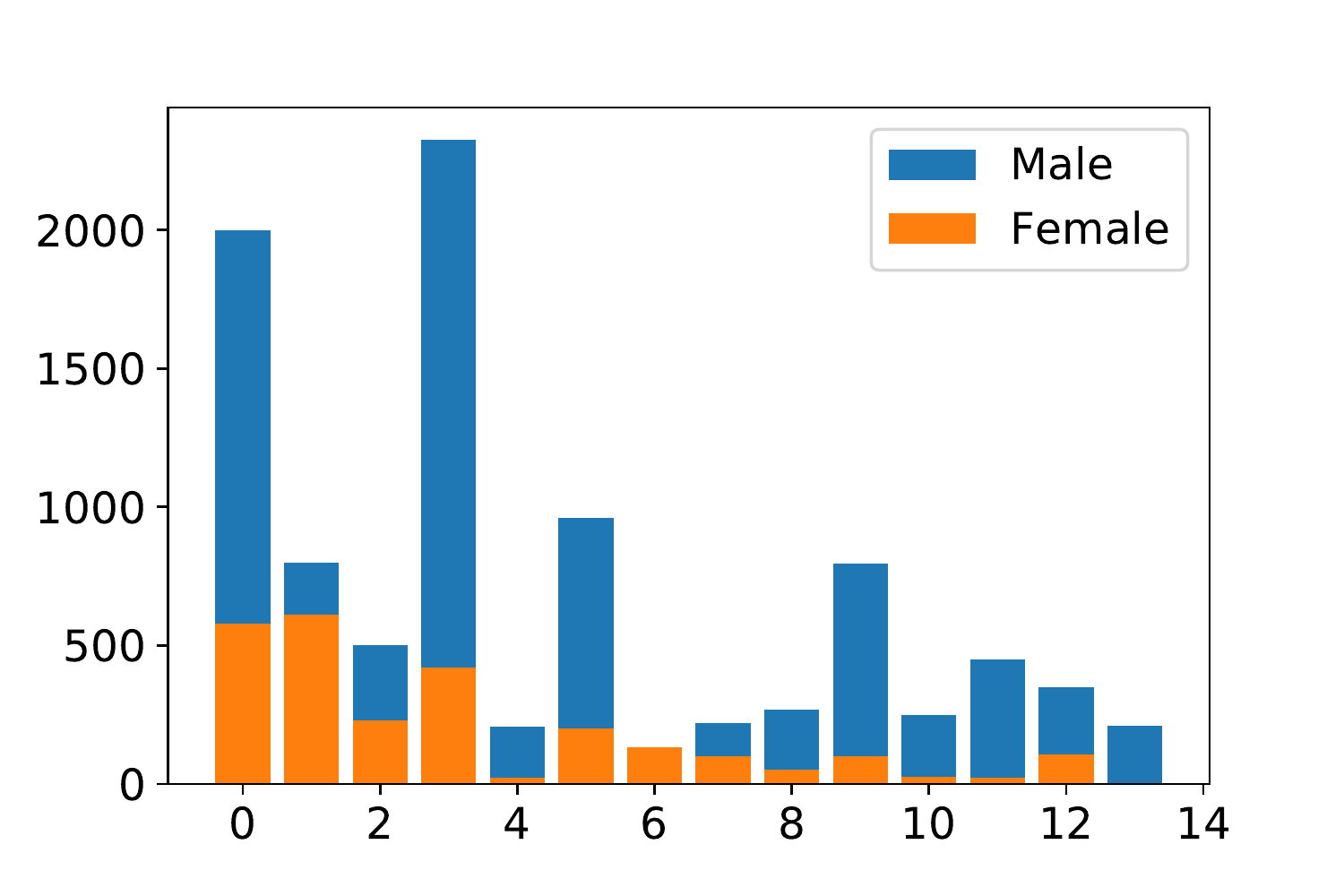}
        \caption{
        DE
        }
        \label{fig:de_stat}
    \end{subfigure}
    ~
    \begin{subfigure}[b]{0.23\textwidth}
        \includegraphics[width=1.18\textwidth]{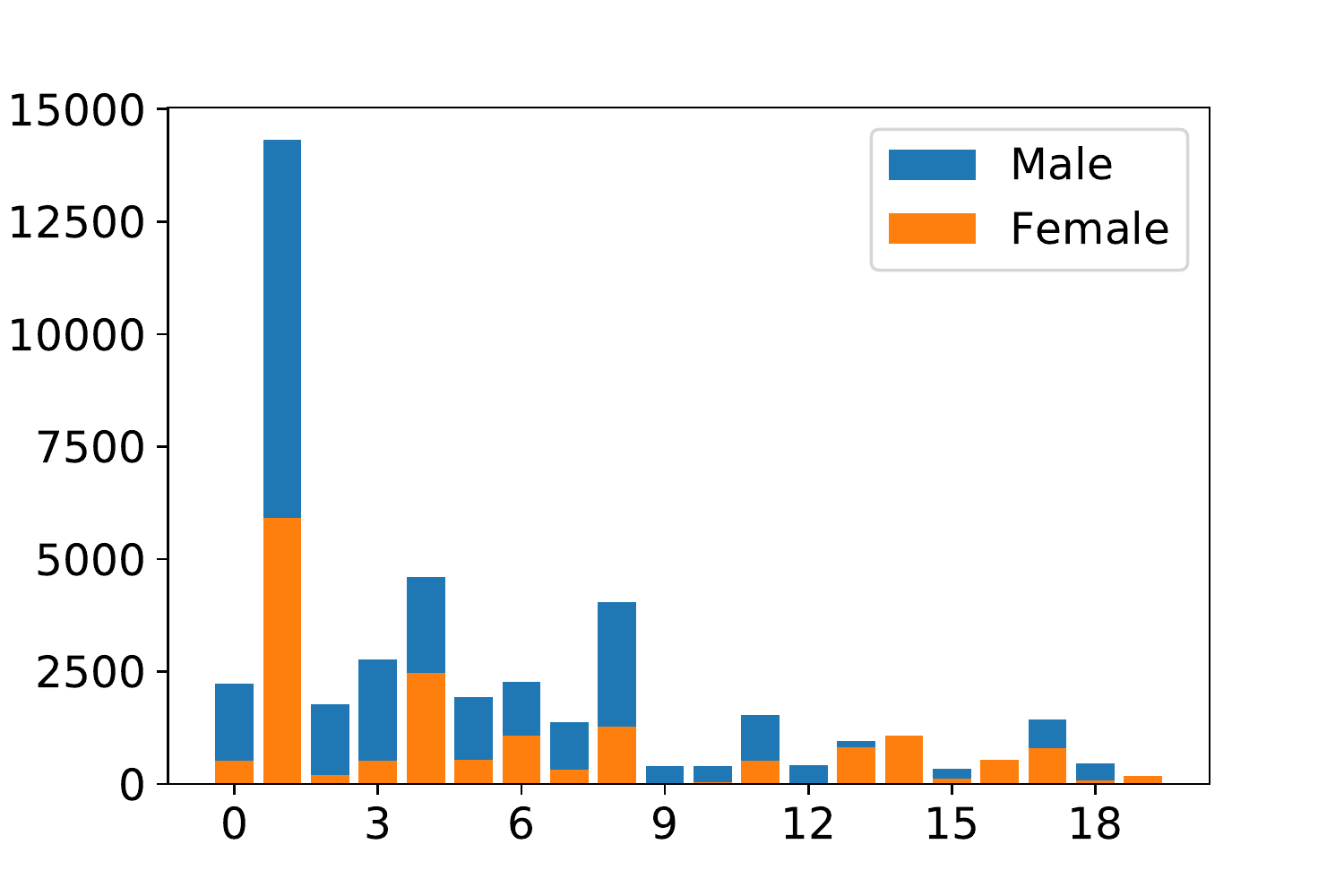}
        \caption{
        FR
        }
        \label{fig:fr_stat}
    \end{subfigure}
    \caption{
    Gender statistics of \mlb~dataset for different occupations where each occupation has at least 200 instances. X-axis here stands for the occupation index and y-axis is the number of instances for each occupation. Among all the languages, EN corpus is the most gender balanced one. All the corresponding occupations will be provided in the appendix.
    }
    \label{fig:bios_data_stat}
\end{figure*}

\subsubsection*{Bias Evaluation} 
We follow the method in ~\citet{zhao2018gender}  to measure the extrinsic bias:  using the performance gap between different gender groups as a metric to evaluate the bias in the \mlb~dataset. We split the dataset based on the gender attribute. A gender-agnostic model  should have similar performance in  each group.  To be specific, we use the average performance gap across each occupation in the male and female groups aggregated across all occupations ($\mid$Diff$\mid$  in Table~\ref{tab:bios_res}) to measure the bias.  However, as described in~\citet{swinger2019biases}, people's names are potentially indicative of their genders. To eliminate the influence of names as well as the gender pronouns on the model predictions, we use a ``scrubbed'' version of the \mlb~dataset by removing the names and some gender indicators (e.g., gendered pronouns and prefixes such as ``Mr.'' or ``Ms.'').

To make  predictions of the occupations, we adopt the model used in ~\citet{de2019bias} by taking the fastText embeddings as the input and encoding the bio text with  bi-directional GRU units following by an attention mechanism. The predictions are generated by a softmax layer. We train such models using standard cross-entropy loss and keep the embeddings frozen during the training.

\subsection{Characterizing Bias in Multilingual Models}

In this section, we analyze the bias in the multilingual word embeddings from the extrinsic perspective. We show that bias exists in  cross-lingual transfer learning and the bias in multilingual word embeddings contributes to such bias.

The gender distribution of the ~\mlb~dataset is shown in Fig.~\ref{fig:bios_data_stat}. Among the three languages, EN corpus is most gender neutral one where the ratio between male and female instances is around $1.2:1$. For all the other languages, male instances are far larger than female ones. In ES, the ratio between male and female is $2.7:1$, in DE it is $3.53:1$,  and in FR, it is $2.5:1$; all are biased towards the  male gender.

\begin{table}[t]
\small
\centering
\begin{adjustbox}{max width=\textwidth}
    \begin{tabular}{|c|c|c|c|c|c|}
        \hline
          \mlb &  Emb. &  Avg. & Female &  Male & $\mid$Diff$\mid$ \\
          \hline
          \multirow{5}{*}{EN} & en & 82.82 & 84.69 & 80.70 & \textbf{7.26} \\
          \cline{2-6}
          & endeb & 83.00 & 84.71 & 81.06 & 6.09 $\downarrow$ \\
          \cline{2-6}
           & en-es & 83.43 & 85.14  & 81.51  & 6.72 $\downarrow$ \\
        \cline{2-6}
           & en-de & 82.85  & 84.64 & 80.84  & 6.37 $\downarrow$  \\
           \cline{2-6}
          & en-fr & 82.66 & 84.34 & 80.78 & 5.87 $\downarrow$ \\
           \hline
           \hline
           \multirow{5}{*}{ES} & es &  63.83 & 64.47 & 63.56 & 6.56 \\
          \cline{2-6}
          & es-en & 61.47 & 61.42 & 61.49  & \textbf{7.13} $\uparrow$\\
          \cline{2-6}
           & es-endeb & 61.91 & 62.98 & 61.45   & 5.61 $\downarrow$ \\
        \cline{2-6}
           & es-de & 61.61 & 62.82 & 61.11 & 5.51  $\downarrow$ \\
           \cline{2-6}
          & es-fr & 62.91 & 63.31 & 62.73 & 4.32 $\downarrow$ \\
           \hline
  \end{tabular}
  \end{adjustbox}
    \caption{Results on scrubbed \mlb.  ``Emb.'' stands for the embeddings used in  model training. ``Avg.'', ``Female'' and ``Male''  refer to the overall average accuracy (\%), and average accuracy for different genders respectively. `` $\mid$Diff$\mid$''  stands for the average absolute accuracy gap between each occupation in the male and female groups aggregated across all the occupations. The results of FR and DE are in the appendix.}
    \label{tab:bios_res}
\end{table}

\paragraph{Bias in Monolingual BiosBias} 
We first evaluate the bias in the \mlb~monolingual dataset by predicting the occupations of the bios in each language.\footnote{The results of DE and FR are  in the appendix.} From Table~\ref{tab:bios_res} we observe that: 1) Bias commonly exists across all languages  ($\mid$Diff$\mid$ $>0$) when using different aligned embeddings, meaning that the model works differently for male and female groups.  2) When training the model using different aligned embeddings, it does not affect the overall average performance significantly (``Avg.'' column in the table). 3) The alignment direction influences the bias. On training the model based on the embeddings aligned to different target space,  we find that aligning the embeddings to ENDEB or a gender-rich language reduces the bias in the downstream task. This is aligned with our previous observation in Section~\ref{sec:intrinsic}.

\begin{table}[!t]
\small
\centering
    \begin{tabular}{|@{ }c@{ }|c@{ }|@{ }c@{ }|c|@{ }c@{ }|c|c|}
        \hline
        Trans. &  Src. & Tgt. & Avg. & Female & Male & $\mid$Diff$\mid$ \\
          \hline
          
        \multirow{2}{*}{EN$\rightarrow$ES} & en & es-en & 41.68 & 42.29 & 41.42 &2.83  \\
         \cline{2-7}
          & en-es & es & 34.15 & 33.97 & 34.22 & 3.49 \\
         \hline
        \hline
        \multirow{2}{*}{ES$\rightarrow$EN} & es & en-es & 57.33 & 59.61 & 54.75 & 8.33 \\
         \cline{2-7}
        & es-en & en & 57.05 & 59.32 & 54.47 & 10.13 \\
        \hline
         
  \end{tabular}
    \caption{Results of transfer learning on the scrubbed \mlb. ``Src.'' and ``Tgt.'' stand for the embeddings in source model and fine tuning procedure respectively. }
    \label{tab:bios_transfer_res}
\end{table}

\begin{table}[t]
\small
\centering
    \begin{tabular}{|@{ }c@{ }|c@{ }|@{ }c@{ }|c|@{ }c@{ }|c|c|}
        \hline
        Trans. &   Src. & Tgt. & Avg. & Female & Male & $\mid$Diff$\mid$ \\
          \hline
          
        \multirow{4}{*}{EN$\rightarrow$ES} & en & es-en & 39.17  &  41.30 & 38.70 &  \textbf{7.97} \\
        \cline{2-7}
        & en-es & es & 35.66 & 36.11 & 35.47 & 4.53 \\
         \cline{2-7}
        &  en-de & es-de & 34.12 & 34.46  & 33.98   & 4.07   \\
        \cline{2-7}
        & en-fr & es-fr & 37.63 & 38.75 & 37.16 & 4.87 \\
        \hline
        \hline
        \multirow{4}{*}{ES$\rightarrow$EN} & es & en-es & 58.41 & 61.78 & 54.60 & 9.03 \\
         \cline{2-7}
         & es-en & en & 55.62 & 58.00 & 52.93 & \textbf{9.52} \\
         \cline{2-7}
         & es-de & en-de & 57.98 & 60.47 & 55.17 & 9.13 \\
         \cline{2-7}
         & es-fr & en-fr & 55.04 & 57.85 & 51.86 & 8.47 \\
        
        \hline
  \end{tabular}
    \caption{Results of transfer learning on gender balanced scrubbed \mlb. The bias in the last column demonstrates that the bias in the multilingual word embeddings  also influences  bias in  transfer learning.}
    \label{tab:bios_multilingual_bias}
\end{table}

\begin{table}[!t]
\small
\centering
    \begin{tabular}{|@{ }c@{ }|@{ }c@{ }|@{ }c@{ }|@{ }c@{ }|@{ }c@{ }|@{ }c@{ }|@{ }c@{ }|}
        \hline
        Trans. &   Src. & Tgt. & Avg. & Female & Male & $\mid$Diff$\mid$ \\
          \hline
          
        \multirow{1}{*}{EN$\rightarrow$ES} & endeb & es-endeb & 37.44 & 39.90 & 36.40 & 5.93  \\
        \hline
        \multirow{1}{*}{ES$\rightarrow$EN} & es-endeb & endeb &  52.51& 54.45& 50.03& 9.06 \\
        \hline
         
  \end{tabular}
    \caption{Bias mitigation results of transfer learning when we aligned the embeddings to the ENDEB space on gender balanced scrubbed \mlb. }
    \label{tab:bios_reduce_bias}
\end{table}

\paragraph{Bias in  Transfer Learning}
Multilingual word embeddings are widely used in cross-lingual transfer learning~\cite{ruder2017survey}. In this section, we conduct experiments to understand how the  bias in multilingual word embeddings  impacts the bias in  transfer learning. 
To do this, we train our model in one language  (i.e., source language) and transfer it to another language  based on the aligned embeddings obtained in Section~\ref{sec:multi_emb}. 
For the transfer learning, we  train the model on the training corpus of the source language and randomly choose 20\% of the dataset from the target  language and use them to fine-tune the model.\footnote{As there are fewer examples in DE, we  use the whole datasets for transfer learning.} Here, we do not aim at achieving state-of-the-art transfer learning performance but pay more attention to the bias analysis.
 Table~\ref{tab:bios_transfer_res}   shows that the bias is  present when we do the transfer learning regardless of the direction of  transfer learning.

\paragraph{Bias from Multilingual Word Embeddings}
The transfer learning bias in Table~\ref{tab:bios_transfer_res} is a combined consequence of both corpus bias and the multilingual word embedding bias. To better understand the influence of the bias in multilingual word embeddings on the transfer learning, we make the training corpus gender balanced for each occupation by upsampling to approximately make the model free of the corpus bias. We then test the bias for different languages with differently aligned embeddings. 
The results are shown in Table~\ref{tab:bios_multilingual_bias}. When we adopt the embeddings aligned to gender-rich languages, we could reduce the bias in the transfer learning, whereas adopting the embeddings aligned to EN results in an increased bias.

\paragraph{Bias after Mitigation}
Inspired by the method in \citet{zhao2018gender}, we mitigate the bias in the downstream tasks by adopting the bias-mitigated word embeddings. To get the less biased multilingual word embeddings, we align other embeddings to the ENDEB space previously obtained in Section~\ref{sec:intrinsic}. Table~\ref{tab:bios_reduce_bias} demonstrates that by adopting such less biased  embeddings, we can  reduce the bias in transfer learning. Comparing to  Table~\ref{tab:bios_multilingual_bias}, aligning the embeddings to a gender-rich language achieves better bias mitigation  and, at the same time, remains the overall performance.

\subsection{Bias Analysis Using Contextualized Embeddings} 

\begin{table}[!t]
\small
\centering
    \begin{tabular}{|c|c|c|c|c|}
    \hline
    \mlb & Avg. & Female & Male & $\mid$Diff$\mid$    \\ 
    \hline
    EN  & 84.35 &85.54 & 83.01 & 7.31 \\
    \hline
    ES & 67.93 & 65.79 & 68.82  &  4.16 \\
    \hline
    DE & 72.68 & 73.68 & 72.28  &  4.89\\
    \hline
     FR & 79.18 & 78.80 & 79.35 &  8.75  \\
    \hline
   
  \end{tabular}
  \caption{Bias in monolingual \mlb  using M-BERT.}
    \label{tab:bios_mono_bert}
  \end{table}

\begin{table}[!t]
\small
\centering
    \begin{tabular}{|@{ }c@{ }|c|c|c|c|}
    \hline
    Trans. & Avg. & Female & Male & $\mid$Diff$\mid$    \\ 
    \hline
    EN$\rightarrow$ES & 66.56  & 65.70 & 66.92  & 5.48  \\
    \hline
     EN$\rightarrow$DE & 76.21 & 75.66 & 76.42 & 7.51\\
    \hline
    EN$\rightarrow$FR & 76.46  &  75.73 &  76.81 &  8.97 \\
    \hline
  \end{tabular}
  \caption{Bias in \mlb  using M-BERT when transferring from EN to other languages. Comparing to multilingual word embeddings, M-BERT achieves better transfer  performance on the \mlb dataset across different languages. But the bias can be higher comparing to the multilingual word embeddings.}
    \label{tab:bios_bert}
  \end{table}

Contextualized embeddings such as ELMo~\cite{peters2018elmo}, BERT~\cite{devlin2018bert} and XLNet~\cite{yang2019xlnet} have shown significant performance improvement in various NLP applications. Multilingual BERT (M-BERT) has shown its great ability for the transfer learning. 
As M-BERT provides one single language model trained on multiple languages, there is no longer a need for alignment procedure. 
In this section, we analyze the bias in monolingual \mlb~dataset as well as in transfer learning  by replacing the fastText embeddings with M-BERT embeddings.
Similar to previous experiments, we train the model on the English dataset and transfer to other languages. 
Table \ref{tab:bios_mono_bert} and \ref{tab:bios_bert} summarizes our results: comparing to results by fastText embeddings in Table~\ref{tab:bios_res}, M-BERT  improves the performance on monolingual \mlb~dataset as well as the transfer learning tasks. When it comes to the bias, using M-BERT  gets similar or lower bias in the monolingual datasets, but sometimes achieves higher bias than the multilingual word embeddings in transfer learning tasks such as the EN $\rightarrow$ ES (in Table~\ref{tab:bios_transfer_res}).

\section{Conclusion}
\label{sec:conclusion}
Recently bias in embeddings has attracted much attention. However, most of the work   only focuses on English corpora and little is known about the bias in multilingual embeddings. In this work, we build different metrics and datasets to analyze gender bias in the multilingual embeddings from both the intrinsic and extrinsic perspectives. We show that gender bias commonly exists across different languages and the alignment target for generating multilingual word embeddings also affects such bias.
In practice, we can choose the embeddings aligned to a gender-rich language to reduce the bias.  

However, due to the limitation of  available resources, this study is limited to the European languages.
We hope this study can work as a foundation to motivate future research about the analysis and mitigation of bias in  multilingual embeddings.
We encourage researchers to look at languages with different grammatical gender (such as Czech and Slovak) and propose new methods to reduce the bias in multilingual embeddings as well as in cross-lingual transfer learning.


\section*{Acknowledgments}
This work was supported in part by NSF Grant IIS-1927554. We would like to thank Maria De-Arteaga and Andi Peng for the helpful discussion, and thank all the reviewers for their feedback.

\bibliography{acl2020,uclanlp}
\bibliographystyle{acl_natbib}

\appendix
\section{Appendices}
\label{sec:appendix}

\subsection{Multilingual Word Embeddings Alignment}

We use the default hyperparameters in the RCSLS alignment model (\url{https://github.com/facebookresearch/fastText}) but change batch size to 5000 and set ``sgd'' to true to make sure the batch size is used. The ``maxsup'' is set to  the same as ``maxneg'' with 200000. 

\subsection{Intrinsic Bias Analysis}
\begin{table}[!h]
\small
\centering
    \begin{tabular}{|c|c|c|c|c|}
        \hline
          \multirow{2}{*}{Category} & \multicolumn{4}{c|}{Target} \\
          \cline{2-5}
           & EN & ES & DE & FR \\
           \hline
           Strong-gendered &  0.1138 & 0.0848 &  0.0935 & 0.0833 \\
           \hline
           Weak-gendered & 0.0477 & 0.0400 & 0.0430 &  0.0395\\
           \hline
    \end{tabular}
    \caption{Bias in EN before and after alignment to different languages for different word categories. For different situation, again we see the bias will reduce when we align the words to gender rich languages.} 
    \label{tab:bias_res_cats}
\end{table}

\begin{table}[!h]
\small
\centering
    \begin{tabular}{|c|c|c|c|c|}
        \hline
          \multirow{2}{*}{Category} & \multicolumn{4}{c|}{Target} \\
          \cline{2-5}
           & ENDEB & ES & DE & FR \\
           \hline
           Strong-gendered &  0.0830 & 0.0683 &  0.0747 & 0.0685 \\
           \hline
           Weak-gendered & 0.0126 & 0.0201 & 0.0269 &  0.0162\\
           \hline
    \end{tabular}
    \caption{Bias in ENDEB before and after alignment to different languages for different word categories. When aligning to a gender rich language, the bias in those strong-gendered words reduces.} 
    \label{tab:bias_res_endeb_cats}
\end{table}

\subsection{Transfer Learning Setting}
For the transfer learning, we filter some occupations that commonly occur across all languages and manually make the distribution of each occupation similar in each language. For each corpus, we use 60\% of the corpus for training, 20\% for validation and 20\% for testing.

\subsection{Occupation Lists for \mlb~ Gender Statistics}
We list all the occupations for each language in Fig.~\ref{fig:bios_data_stat}.

\textbf{EN}: professor, accountant, journalist, architect, photographer, psychologist, teacher, nurse, attorney, software\_engineer, painter, physician, chiropractor, personal\_trainer, surgeon, filmmaker, dietitian, dentist, dj, model, composer, poet, comedian, yoga\_teacher, interior\_designer, pastor, rapper, paralegal
\\
\textbf{ES}: student, model, teacher, cook, musician, artist, painter, professor, administrator, scientist, writer, nurse, hotelier, lawyer, coach, computer\_programmer, doctor, journalist, architect, soldier, pharmacist, poet, dancer, engineer, farmer, pianist, pilot, psychologist, surgeon, athlete, mechanic, driver, accountant, rapper, photographer, filmmaker, attorney, physician, dj, comedian, composer\\
\textbf{DE}: journalist, teacher, psychologist, attorney, dj, photographer, nurse, professor, pastor, architect, filmmaker, composer, painter, software\_engineer\\
\textbf{FR}: filmmaker, teacher, composer, painter, journalist, physician, attorney, poet, photographer, pastor, rapper, architect, dj, comedian, psychologist, accountant, nurse, model, surgeon, dietitian

\begin{table}[!t]
\small
\centering
    \begin{tabular}{|c|c|c|c|c|c|}
        \hline
          \mlb &  Emb. &  Avg. & Female &  Male & $\mid$Diff$\mid$ \\
          \hline
          \multirow{5}{*}{DE} & de &  55.4  &  59.87  & 53.63  &  10.42 \\
          \cline{2-6}
          & de-en & 56.88  & 61.84  &  54.92  & \textbf{15.41} \\
          \cline{2-6}
           & de-endeb & 54.09 & 55.26   & 53.63    &  6.54 \\
        \cline{2-6}
           & de-es & 54.46 &  56.58 & 53.63 & 9.51  \\
           \cline{2-6}
          & de-fr & 55.8 & 57.50 & 55.18 & 10.43  \\
           \hline
            \multirow{5}{*}{FR} & fr & 76.52  & 76.24  & 76.65 & 11.58  \\
          \cline{2-6}
          & fr-en &  74.13 & 74.87 &  73.79  & \textbf{12.96} \\
          \cline{2-6}
           & fr-endeb & 73.92 & 74.19   & 73.79     &  10.84 \\
        \cline{2-6}
           & fr-es & 74.57 &  74.19 & 74.74 & 11.23   \\
           \cline{2-6}
          & fr-de & 75.11 & 75.56 & 74.90 & 12.07  \\
           \hline

  \end{tabular}
    \caption{Results on the scrubbed BiosBias dataset in DE and FR. }
    \label{tab:bios_fr_res}
\end{table}

\begin{table}[!t]
\small
\centering
    \begin{tabular}{|@{ }c@{ }|c@{ }|@{ }c@{ }|c|@{ }c@{ }|c|c|}
        \hline
        Trans. &  Src. & Tgt. & Avg. & Female & Male & $\mid$Diff$\mid$ \\
          \hline
          
        \multirow{2}{*}{EN$\rightarrow$DE} & en & de-en & 37.55   & 39.47  & 36.79  & 16.52   \\
         \cline{2-7}
        &  en-de & es-de & 34.57 &32.89 & 35.23 & 13.58\\
        \hline
        \multirow{2}{*}{DE$\rightarrow$EN} & de & en-de & 42.47 & 45.76  & 38.77  & 6.46 \\
         \cline{2-7}
         & de-en & en &  38.55&  41.25& 35.51& 7.12 \\
        \hline
  \end{tabular}
    \caption{Results of transfer learning  between EN and DE on \mlb~dataset.}
    \label{tab:bios_transfer_de_res}
\end{table}

\subsection{Extrinsic Bias Results in DE and FR}
We show the bias in monolingual DE and FR datasets in Table~\ref{tab:bios_fr_res} and in the transfer learning between EN and them in Table~\ref{tab:bios_transfer_de_res} and~\ref{tab:bios_transfer_fr_res} respectively.

\begin{table}[!t]
\small
\centering
\resizebox{\columnwidth}{!}{
    \begin{tabular}{|@{ }c@{ }|c@{ }|@{ }c@{ }|c|@{ }c@{ }|c|c|}
        \hline
        Trans. &  Src. & Tgt. & Avg. & Female & Male & $\mid$Diff$\mid$ \\
          \hline
          
        \multirow{2}{*}{EN$\rightarrow$FR} & en & fr-en &  41.43 & 41.03 &  41.62 & 5.96   \\
         \cline{2-7}
         & en-fr & en & 43.12& 44.96 & 42.26 & 8.33 \\
       
        \hline
        \multirow{2}{*}{FR$\rightarrow$EN} & fr & en-fr & 57.81 & 62.02  & 51.94  &  9.79 \\
         \cline{2-7}
          &  fr-en & fr & 55.15 & 58.83  & 50.0  &  8.3 \\
        \hline
  \end{tabular}
  }
    \caption{Results of transfer learning  between EN and FR on \mlb~dataset.}
    \label{tab:bios_transfer_fr_res}
\end{table}

Table~\ref{tab:bios_balanced_de_bias} and~\ref{tab:bios_balanced_fr_bias} is the bias result of the transfer learning between EN and DE, FR when we manually make the gender ratio balanced for each occupation in the corpus. We also show the mitigation results when we align all the embeddings to the ENDEB space.

\begin{table}[!h]
\small
\centering
    \begin{tabular}{|@{ }c@{ }|@{}c@{ }|@{ }c@{ }|@{ }c@{ }|@{ }c@{ }|@{ }c@{ }|@{ }c@{ }|}
        \hline
        Trans. &   Src. & Tgt. & Avg. & Female & Male &$\mid$Diff$\mid$ \\
          \hline
          
        \multirow{5}{*}{EN$\rightarrow$DE} & en & de-en & 39.40  & 38.28  &  39.82 &  10.65 \\
        \cline{2-7}
        & endeb & de-endeb & 33.51 & 31.37& 34.42&  8.9\\
         \cline{2-7}
        & en-es & de-es & 33.16 & 32.21  & 33.50 &  9.31 \\
         \cline{2-7}
        &  en-de & de & 33.96 & 31.02  & 35.03  & 9.13   \\
        \cline{2-7}
        & en-fr & de-fr & 38.31 & 34.17 & 39.82 & \textbf{11.04} \\
       
        \hline
        \multirow{5}{*}{DE$\rightarrow$EN} & de & en-de & 46.43  & 48.83  & 43.72 & 7.93  \\
         \cline{2-7}
         & de-en & en & 50.48 & 53.91 & 46.58 & \textbf{8.10} \\
         \cline{2-7}
         & de-endeb & endeb & 44.44 & 46.84& 41.73 & 7.16 \\
         \cline{2-7}
         & de-es & en-es & 44.04 & 47.54 &40.09 &  7.29 \\
         \cline{2-7}
         & de-fr & en-fr & 46.01 & 47.57 & 44.25 & 7.03 \\
         
        \hline
  \end{tabular}
    \caption{Results of transfer learning between EN and DE on the scrubbed BiosBias dataset when we make the dataset gender balanced. The bias in the last column demonstrates that the bias in the multilingual word embeddings will also influence the bias in the transfer learning.}
    \label{tab:bios_balanced_de_bias}
\end{table}

\begin{table}[!h]
\small
\centering
    \begin{tabular}{|@{ }c@{ }|c@{ }|@{ }c@{ }|@{ }c@{ }|@{ }c@{ }|@{ }c@{ }|@{ }c@{ }|}
        \hline
        Trans. &   Src. & Tgt. & Avg. & Female & Male & $\mid$Diff$\mid$ \\
          \hline
          
        \multirow{5}{*}{EN$\rightarrow$FR} & en & fr-en & 36.66  & 36.24 & 36.85 &  \textbf{7.97} \\
         \cline{2-7}
          & endeb & fr-endeb & 34.86 & 32.82&  35.82& 5.44\\
       
        \cline{2-7}
         & en-es & fr-es & 34.82 & 34.19 & 35.11 &  6.77 \\
       
        \cline{2-7}
        &  en-de & fr-de & 33.51 & 33.85  & 33.36   & 5.78   \\
        \cline{2-7}
        & en-fr & fr & 35.68 & 33.50 & 36.70 & 6.81 \\
        \hline
        \multirow{5}{*}{FR$\rightarrow$EN} & fr & en-fr & 59.21 & 61.55 & 55.94 & 10.3 \\
         \cline{2-7}
          & fr-en & en & 50.80 & 54.44 & 45.73 & \textbf{11.42} \\
         \cline{2-7}
         & fr-endeb & endeb & 49.33 & 52.91 & 44.33 & 10.14 \\
        
         \cline{2-7}
         & fr-es & en-es & 49.28 & 51.86 & 45.66 & 10.42 \\
         \cline{2-7}
         & fr-de & en-de & 50.92 & 54.10 & 46.46 & 7.36 \\
        \hline
  \end{tabular}
    \caption{Results of transfer learning between EN and FR on the scrubbed BiosBias dataset when we make the dataset gender balanced. The bias in the last column demonstrates that the bias in the multilingual word embeddings will also influence the bias in the transfer learning.}
    \label{tab:bios_balanced_fr_bias}
\end{table}

\end{document}